\documentclass{article}

\PassOptionsToPackage{numbers, compress}{natbib}

\usepackage[preprint]{neurips_2020}

\usepackage{natbib}
\usepackage{microtype}
\usepackage{graphicx}
\usepackage{subfigure}
\usepackage{booktabs}
\usepackage{makecell}
\usepackage[utf8]{inputenc} %
\usepackage[T1]{fontenc}    %
\usepackage{hyperref} %
\usepackage{url}            %
\usepackage{amsfonts}       %
\usepackage{nicefrac}       %
\usepackage{microtype}      %
\usepackage[font=small]{caption} %
\usepackage{amsmath}        %
\usepackage{amsthm}         %
\usepackage{wrapfig}        %
\usepackage{algorithm}      %
\usepackage{algorithmic}    %
\usepackage{listings}       %
\usepackage{bm}             %
\usepackage{color}
\usepackage{tikz}
\usepackage{tabu}
\usepackage{makecell}
\usepackage{multicol}

\usepackage{amsmath,amsfonts,bm}

\def\Figref#1{Figure~\ref{#1}}

\def\Tabref#1{Table~\ref{#1}}

\def\Secref#1{Section~\ref{#1}}

\def\eqref#1{equation~(\ref{#1})}
\def\Eqref#1{Equation~(\ref{#1})}

\def\Algref#1{Algorithm~\ref{#1}}

\def\1{\bm{1}}

\DeclareMathAlphabet{\mathsfit}{\encodingdefault}{\sfdefault}{m}{sl}
\SetMathAlphabet{\mathsfit}{bold}{\encodingdefault}{\sfdefault}{bx}{n}

\newcommand{\E}{\mathbb{E}}

\newtheorem*{thm*}{Theorem}

\usepackage{colortbl}
\usepackage[frozencache,cachedir=.]{minted}
\usepackage{bbm}
\usepackage{multirow}
\usepackage[export]{adjustbox}
\usepackage[normalem]{ulem}
\usepackage{cleveref}
\definecolor{Gray}{gray}{0.5}
\definecolor{GrayBG}{gray}{0.95}
\usepackage{etoolbox}
\makeatletter
\AfterEndEnvironment{algorithm}{\let\@algcomment\relax}
\AtEndEnvironment{algorithm}{\kern2pt\hrule\relax\vskip3pt\@algcomment}
\let\@algcomment\relax
\newcommand\algcomment[1]{\def\@algcomment{\footnotesize#1}}
\renewcommand\fs@ruled{\def\@fs@cfont{\bfseries}\let\@fs@capt\floatc@ruled
  \def\@fs@pre{\hrule height.8pt depth0pt \kern2pt}%
  \def\@fs@post{}%
  \def\@fs@mid{\kern2pt\hrule\kern2pt}%
  \let\@fs@iftopcapt\iftrue}
\makeatother
\newlength\savewidth

\makeatletter\renewcommand\paragraph{\@startsection{paragraph}{4}{\z@}
  {.5em \@plus1ex \@minus.2ex}{-.5em}{\normalfont\normalsize\bfseries}}\makeatother

\newcommand{\methodname}{{HDGE}}

\newcommand{\titlename}{Hybrid Discriminative-Generative Training via Contrastive Learning}
\title{\titlename{}}

\author{Hao Liu, ~Pieter Abbeel \\
University of California, Berkeley \\
\texttt{\{hao.liu, pabbeel\}@cs.berkeley.edu}
}

\begin{document}

\maketitle

\begin{abstract}

Contrastive learning and supervised learning have both seen significant progress and success.  However, thus far they have largely been treated as two separate objectives, brought together only by having a shared neural network.  In this paper we show that through the perspective of hybrid discriminative-generative training of energy-based models we can make a direct connection between contrastive learning and supervised learning.
Beyond presenting this unified view, we show our specific choice of approximation of the energy-based loss outperforms the existing practice in terms of classification accuracy of WideResNet on CIFAR-10 and CIFAR-100. It also leads to improved performance on robustness, out-of-distribution detection, and calibration. 

\end{abstract}

\section{Introduction}
In the past few years, the field of deep supervised learning has seen significant progress.   Example successes include large-scale image classification~\citep{he2016deep, simonyan2014very, srivastava2015highway, szegedy2016rethinking} on the challenging ImageNet benchmark~\citep{deng2009imagenet}. 
The common objective for solving supervised machine learning problems is to minimize the cross-entropy loss, which is defined as the cross entropy between a target distribution and a categorical distribution called Softmax which is parameterized by the model's real-valued outputs known as logits. The target distribution usually consists of one-hot labels.
There has been a continuing effort on improving upon the cross-entropy loss, various methods have been proposed, motivated by different considerations~\citep{hinton2015distilling,muller2019does,szegedy2016rethinking}.

Recently, contrastive learning has achieved remarkable success in representation learning. Contrastive learning allows learning good representations and enables efficient training on downstream tasks, an incomplete list includes image classification~\citep{chen2020simple, chen2020big, grill2020bootstrap, he2019momentum, tian2019contrastive, oord2018representation}, video understanding~\citep{han2019video}, and knowledge distillation~\citep{tian2019contrastive}.
Many different training approaches have been proposed to learn such representations, usually relying on visual pretext tasks. 
Among them, state-of-the-art contrastive methods~\citep{he2019momentum, chen2020simple, chen2020improved} are trained by reducing the distance between representations of different augmented views of the same image (`positive pairs'), and increasing the distance between representations of augment views from different images (`negative pairs').

Despite the success of the two objectives, they have been treated as two separate objectives, brought together only by having a shared neural network.

In this paper, to show a direct connection between contrastive learning and supervised learning, we consider the energy-based interpretation of models trained with cross-entropy loss, building on~\citet{grathwohl2019your}.
We propose a novel objective that consists of a term for the conditional of the label given the input (the classifier) and a term for the conditional of the input given the label.
We optimize the classifier term the normal way.
Different from~\citet{grathwohl2019your}, we approximately optimize the second conditional over the input with a contrastive learning objective instead of a Monte-Carlo sampling-based approximation.
In doing so, we provide a unified view on existing practice.

Our work takes inspiration from the work by~\citet{ng2002discriminative}.
In their 2002 paper, \citet{ng2002discriminative} showed that classifiers trained with a generative loss (i.e., optimizing $p(x|y)$, with $x$ the input and $y$ the classification label) can outperform classifiers with the same expressiveness trained with a discriminative loss (i.e., optimizing $p(y|x)$).  Later it was shown that hybrid discriminative generative model training can get the best of both worlds~\citep{raina2004classification}.  The work by~\citet{ng2002discriminative} was done in the (simpler) context of Naive Bayes and Logistic Regression.  Our work can be seen as lifting this work into today's context of training deep neural net classifiers.

We extensively evaluate our approach on image classification tasks, out-of-distribution detection, adversarial robustness, and calibration. 
The experiment shows that this novel loss improves classification accuracy of WideResNet-28-10~\citep{zagoruyko2016wide} on CIFAR-10 and CIFAR-100 datasets and outperforms state-of-the-art supervised learning and contrastive learning methods. The method also significantly improves the performance of WideResNet-28-10 on robustness, out-of-distribution detection, and calibration compared with generative models and hybrid models.

\section{Background}

\subsection{Supervised learning}
In supervised learning, given a data distribution $p(x)$ and a label distribution $p(y|x)$ with $C$ categories, a classification problem is typically addressed using a parametric function, $f_\theta: \mathbb{R}^D \rightarrow \mathbb{R}^C$, which maps each data point $x \in \mathbb{R}^D$ to $C$ real-valued numbers termed as logits.
These logits are used to parameterize a categorical distribution using the Softmax function:
\begin{align}
    q_\theta(y|x) = \frac{\exp(f_\theta(x)[y])}{\sum_{y'} \exp(f_\theta(x)[y'])},
    \label{eq:softmax_def}
\end{align}
where $f_\theta(x)[y]$ indicates the $y^{\rm th}$ element of $f_\theta(x)$, $i.e.$, the logit corresponding to the $y^{\rm th}$ class label.
One of the most widely used loss functions for learning $f_\theta$ is minimizing the negative log likelihood:
\begin{align}
  \min_\theta  - \E_{p_{\rm data}(x,y)} \left[ \log q_\theta(y|x) \right]. 
  \label{eq:cross_entropy_loss}
\end{align}
This loss function is often referred to as the cross-entropy loss function, because it corresponds to minimizing the KL-divergence with a target distribution $p(y|x)$, which consists of one-hot vectors with the non-zero element denoting the correct prediction.

\subsection{Energy-based models}
\label{sec:ebm_bg}

\paragraph{Energy-based models.}
Energy based models (EBMs)~\citep{lecun2006tutorial} are based on the observation that probability densities $p(x)$ for $x \in \mathbb{R}^D$ can be expressed as
\begin{align}
    p_\theta(x) = \frac{\exp(-E_\theta(x))}{Z(\theta)}~,
\end{align}
where $E_\theta(x): \mathbb{R}^D \rightarrow \mathbb{R}$ maps each data point to a scalar; and $Z(\theta) = \sum_{x \in \mathcal{X}} \exp(-E_\theta(x))$ (or, for continuous $x$ we'd have $Z(\theta) = \int_{x \in \mathcal{X}} \exp(-E_\theta(x))$) is the normalizing constant, also known as the partition function. Here $\mathcal{X}$ is the full domain of $x$. For example, in the case of (let's say) 16x16 RGB images, computing $Z$ exactly would require a summation over $(256 \times 256 \times 256)^{(16 \times 16)} \approx 10^{2500}$ terms.  

We can parameterize an EBM using any function that takes $x$ as the input and returns a scalar.
For most choices of $E_\theta$, one cannot compute or even reliably estimate $Z(\theta)$, which means estimating the normalized densities is intractable and standard maximum likelihood estimation of the parameters, $\theta$, is not straightforward.

\paragraph{Training EBMs.}
The log-likelihood objective for an EBM consists of a sum of $\log p_\theta(x)$ terms, one term for each data point $x$. The gradient of each term is given by:
\begin{align}
    \frac{\partial \log p_\theta(x)}{\partial \theta}  = \E_{p_\theta(x')}\left[ \frac{\partial E_\theta(x')}{\partial \theta} \right] - \frac{\partial E_\theta(x)}{\partial \theta}~,
    \label{eq:ebm_grad_est}
\end{align}
where the expectation is over the model distribution $p_\theta(x')$.  This expectation is typically intractable (for much the same reasons computing $Z(\theta)$ is typically intractable).  However, it can be approximated through samples--assuming we can sample from $p_\theta$.  Generating exact samples from $p_\theta$ is typically expensive, but there are some well established approximate (sometimes exact in the limit) methods based on MCMC~\citep{grathwohl2019your, du2019implicit, hinton2002training}.

Among such sampling methods, recent success in training (and sampling from) energy-based models often relies on the Stochastic Gradient Langevin Dynamics (SGLD) approach~\citep{welling2011bayesian}, which generates samples by following this stochastic process:
\begin{align}
    x_0 \sim p_0(x), \qquad
    x_{i+1} = x_{i} - \frac{\alpha}{2} \frac{\partial E_\theta(x_i)}{\partial x_i} + \epsilon, \qquad \epsilon \sim \mathcal{N}(0, \alpha)
    \label{eq:sgld}
\end{align}
where $\mathcal{N}(0, \alpha)$ is the normal distribution with mean of $0$ and standard deviation of $\alpha$, and $p_0(x)$ is typically a Uniform distribution over the input domain and the step-size $\alpha$ should be decayed following a polynomial schedule.  The SGLD sampling steps are tractable, assuming the gradient of the energy function can be computed with respect to $x$, which is often the case.  It is worth noting this process does \emph{not} require evaluation the partition function $Z(\theta)$ (or any derivatives thereof).

\paragraph{Joint Energy Models.}
The joint energy based model (JEM)~\citep{grathwohl2019your} shows that classifiers in supervised learning are secretly also energy-based based models on $p(x,y)$.
The key insight is that the logits $f_\theta(x)[y]$ in the supervised cross-entropy loss can be seen as defining an energy-based model over $(x,y)$, as follows:
\begin{align}
    p(x, y) = \frac{\exp (f_\theta(x)[y])}{Z(\theta)},
    \label{eq:jem_xy}
\end{align}
where $Z(\theta)$ is the unknown normalization constant.  I.e., matching this with the typical EBM notation, we have $f_\theta(x)[y] = - E_\theta(x,y)$.
Subsequently, the density model of data points $p(x)$ can be obtained by marginalizing over $y$:
\begin{align}
    p(x) = \frac{\sum_{y} \exp (f_\theta(x)[y])}{Z(\theta)},
    \label{eq:jem_x}
\end{align}
with the energy $E_\theta(x) = - \log \sum_{y} \exp (f_\theta(x)[y])$.
JEM~\citep{grathwohl2019your} adds the marginal log-likelihood $p(x)$ to the training objective, where $p(x)$ is expressed with the energy based model from~\Eqref{eq:jem_x}.  JEM uses SGLD sampling for training.

\subsection{Contrastive Learning}
In contrastive learning~\citep{gutmann2010noise, gutmann2012noise, mnih2013learning, mikolov2013distributed}, it is common to optimize an objective of the following form:
\begin{align}
    \min_\theta -\E_{p_{\rm data}(x)} \left[\log \frac{\exp(h_\theta(x) \cdot h_\theta(x'))}{\sum_{i=1}^K \exp(h_\theta(x) \cdot h_\theta(x_i))}\right],
    \label{eq:contrastive_loss_def}
\end{align}
where $x$ and $x'$ are two different augmented views of the same data point, $h_\theta: \mathbb{R}^D \rightarrow \mathbb{R}^H$ maps each data point to a representation space with dimension $H$. 
The inner product between two vectors is a specific choice of distance metric and is shown to work well in practice~\citep{wu2018unsupervised, salakhutdinov2007learning}. Other distance metrics, such as Euclidean distance, have also been used.

This objective tries to maximally distinguish an input $x_i$ from alternative inputs $x_i'$. 
The intuition is that by doing so, the representation captures important information between similar data points, and therefore might improve performance on downstream tasks.
This is usually called the contrastive learning loss or InfoNCE loss~\citep{oord2018representation} and has achieved remarkable success in representation learning~\citep{grill2020bootstrap, he2019momentum, chen2020improved, tian2019contrastive, khosla2020supervised}.

In practice, using a contrastive loss for representation learning requires either large batch sizes~\citep{chen2020simple} or memory banks~\citep{wu2018unsupervised, he2019momentum, chen2020improved}.
Although the $K$ normalization examples in principle could be any samples~\citep{gutmann2012noise}, in practice, one often uses different augmented samples within the same batch or samples from a memory bank.

In the context of supervised learning, the Supervised Contrastive Loss~\citep{khosla2020supervised} shows that selecting $x_i$ from different categories as negative examples can improve the standard cross-entropy training.
Their objective for learning the representation $h_\theta(x)$ is given by:
\begin{align}
    \min_\theta -\E_{p_{\rm data}(x)} \left[\log \frac{\exp(h_\theta(x) \cdot h_\theta(x'))}{\sum_{i=1}^K \exp(h_\theta(x) \cdot h_\theta(x_i)) \mathbbm{1}[y \neq y_i]} \right].
    \label{eq:scl_obj}
\end{align}
We'll see that our approach outperforms Supervised Contrastive Learning from~\citep{khosla2020supervised}, while also simplifying by removing the need for selecting negative examples or pre-training a representation.  Through the simplification we might get a closer hint at where the leverage is coming from.

\section{Hybrid Discriminative Generative Energy-based Model (HDGE)}
\label{sec:method}

As in the typical classification setting, we assume we are given a dataset $(x,y) \sim p_{\rm data}$.  The primary goal is to train a model that can classify ($x$ to $y$).  In addition, we would like the learned model to be capable of out-of-distribution detection, providing calibrated outputs, and serving as a generative model.

To achieve these goals, we propose to train a hybrid model, which consists of a discriminative conditional and a generative conditional by maximizing the sum of both conditional log-likelihoods:

\begin{align}
    \min_\theta - \E_{p_{\rm data}(x, y)} \left[\log q_\theta(y|x) + \log q_\theta(x|y) \right],\label{eq:HDGE}
\end{align}
where $q_\theta(y|x)$ is a standard Softmax neural net classifier, and where
    $q_\theta(x|y) = \frac{\exp(f_\theta(x)[y])}{Z(\theta)}$,
with $Z(\theta) = \sum_x \exp(f_\theta(x)[y])$.%

The rationale for this objective originates from~\citep{ng2002discriminative, raina2004classification}, where they discuss the connections between logistic regression and naive Bayes, and show that hybrid discriminative and generative models can out-perform purely generative or purely discriminative counterparts.

The main challenge with the objective from \Eqref{eq:HDGE} is the intractable partition function $Z(\theta)$.

Our main contribution is to propose a (crude, yet experimentally effective) approximation with a contrastive loss:
\begin{align}
  & \E_{p_{\rm data}(x,y)} \left[ \log q_\theta(x | y) \right]\\
  & = \E_{p_{\rm data}(x,y)} \left[ \log \frac{\exp(f_\theta(x)[y])}{Z(\theta)} \right] \\ 
  & \approx \E_{p_{\rm data}(x,y)}  \left[\log \frac{ \exp(f_\theta(x)[y])}{\sum_{i=1}^K \exp(f_\theta(x_i)[y])} \right ],
  \label{eq:k_approx}
\end{align}
where $K$ denotes the number of normalization samples. 
This is similar to existing contrastive learning objectives, although in our formulation, we also use labels.

Intuitively, in order to have an accurate approximation in~\Eqref{eq:k_approx}, $K$ has to be sufficiently large---becoming exact in the limit of summing over all $x \in \mathcal{X}$.  We don't know of any formal guarantees for our proposed approximation, and ultimately the justification has to come from our experiments.  Nevertheless, there are two main intuitions we considered: (i) We try to make $K$ as large as is practical.  Increasing $K$ is not trivial as it requires a larger memory.  To still push the limits, following \citet{he2019momentum} we use a memory bank to store negative examples.  More specifically, we resort to using a queue to store past logits, and sample normalization examples from this queue during the training process.  (ii) While in principle we would need to sum over all possible $x \in \mathcal{X}$, we could expect to achieve a good approximation by focusing on $(x,y)$ that have low energy.  Since the training examples $x_i$ are encouraged to have low energy, we draw from those for our approximation.  It is worth noting that the training examples $x_i, y_i$ are getting incorporated in the denominator using the same label $y$ as in the numerator.  So effectively this objective is (largely) contrasting the logit value $f_\theta(x)[y]$ for $x$ with label $y$ from the logit values of other training examples $x_i$ that don't have the same label $y$.

To bring it all together, our objective can be seen as a hybrid combination of supervised learning and contrastive learning given by:
\begin{align}
&    \min_\theta - \E_{p_{\rm data}(x, y)} \left[ \alpha \log q_\theta(y|x) + (1-\alpha) \log q_\theta(x|y) \right] \\
& \approx    \min_\theta - \E_{p_{\rm data}(x,y)} \left[\alpha \log \frac{ \exp(f_\theta(x)[y])}{\sum_{y'} \exp(f_\theta(x)[y'])} + (1-\alpha) \log \frac{ \exp(f_\theta(x)[y])}{\sum_{i=1}^K \exp(f_\theta(x_i)[y])} \right ],
    \label{eq:hybrid_obj}
\end{align}
where $\alpha$ is weight between $[0, 1]$. 
When $\alpha=1$, the objective reduces to the standard cross-entropy loss, while $\alpha=0$, it reduces to an end-to-end supervised version of contrastive learning. 
We evaluated these variants in experiments, and we found that $\alpha=0.5$ delivers the highest performance on classification accuracy as well as robustness, calibration, and out-of-distribution detection.

The resulting model, dubbed Hybrid Discriminative Generative Energy-based Model~(\methodname{}), learns to jointly optimize supervised learning and contrastive learning.
The corresponding pseudo code in PyTorch-like~\citep{paszke2019pytorch} style is in~\Algref{alg:code}.
\begin{algorithm}[!htbp]
\caption{Pseudocode of \methodname{} in a PyTorch-like style.
}
\label{alg:code}
\algcomment{
\fontsize{7.2pt}{0em}\selectfont
\texttt{einsum}: Einstein sum; \texttt{cat}: concatenation; \texttt{logsumexp}: LogSumExp operation.
}
\begin{multicols}{2}
\begin{minted}[breaklines, breakanywhere, fontsize=\fontsize{9.5pt}{9.5pt}\selectfont, baselinestretch=0.5, breaksymbolleft=]{python}
# f: encoder networks for images
# queue: queue of probabilities (KxC)
# B: batch size; K: queue size; C: number of classes; m: momentum; t: temperature

for x, y in loader:  # load a minibatch with B images x and labels y
    x = aug(x)  # optionally randomly data augmentation
    y = t(y) # transform to one-hot vector with dimension C
    
    # logits: KxC
    ce_logits = f.forward(x)  

    # standard cross-entropy loss
    # optimize log q(y|x)
    ce_loss = CrossEntropyLoss(ce_logits, y)
    
    # normalize logits by L2 or softmax: KxC
    probs = normalize(ce_logits) 

    # positive logits: Bx1
    l_pos = logsumexp(probs*y, dim=1, keepdim=True)

    # negative logits: BxK
    l_neg = einsum("nc,ck->nck", [y, queue.clone().detach()])  # BxCxK
    l_neg = logsumexp(l_neg, dim=1)  # BxK

    # logits: Bx(1+K)
    logits = cat([l_pos, l_neg], dim=1)

    # positives are the 0-th
    labels = zeros(K)  
    
    # contrastive loss
    # optimize log q(x|y)
    cl_loss = CrossEntropyLoss(logits/t, labels)
    
    # hybrid training
    loss = ce_loss + cl_loss

    # SGD update: encoder network
    loss.backward()
    update(f.params)

    # update buffer
    enqueue(queue, probs)  # enqueue the current minibatch of probs
    dequeue(queue)  # dequeue the earliest minibatch
\end{minted}
\end{multicols}
\end{algorithm}

\section{Experiment}
\label{sec:experiment}
In our experiments, we are interested in answering the following questions:
\begin{itemize}
    \item Does \methodname{} improve classification accuracy over the state-of-the-art supervised learning loss? (\Secref{sec:image_clc})
    \item Does \methodname{} improve the out-of-distribution detection? (\Secref{sec:ood_exp})
    \item Does \methodname{} add better calibration to the model? (\Secref{sec:cal_exp})
    \item Does \methodname{} lead to a model with better adversarial robustness? (\Secref{sec:adv_robust})
    \item How much does the number of negative examples affect \methodname{} performance? (\Secref{sec:good_approximation})
    \item Is \methodname{} able to improve the generative modeling performance? (\Secref{sec:gen_model})
\end{itemize}
To do so, we completed a thorough empirical investigation on several standard datasets: CIFAR-10 and CIFAR-100~\citep{krizhevsky2009learning}, two labeled datasets composed of $32\times32$ images with $10$ and $100$ classes respectively (\Cref{sec:image_clc,,sec:ood_exp,,sec:cal_exp,,sec:gen_model}); SVHN~\citep{netzer2011reading}, a labeled dataset composed of over $600,000$ digit images (\Cref{sec:ood_exp}); CelebA~\citep{liu2015deep}, a labeled dataset consisting of over $200,000$ face images and each with $40$ attribute annotation (\Cref{sec:ood_exp}).

Our experiments are based on WideResNet-28-10~\citep{zagoruyko2016wide}, where we compare \methodname{} against the state-of-the-art cross-entropy loss, against the hybrid method JEM~\citep{grathwohl2019your}, and against the Supervised Contrastive Loss~\citep{khosla2020supervised}. We will use $K=65536$ throughout the experiments if not otherwise mentioned and temperature $\tau=0.1$. The cross-entropy baseline is based on the code from the official PyTorch training code~\footnote{https://github.com/szagoruyko/wide-residual-networks/tree/master/pytorch}.~\methodname{}'s implementation is based on the official codes of MoCo~\footnote{https://github.com/facebookresearch/moco} and JEM~\footnote{https://github.com/wgrathwohl/JEM}. 
Our source code is available online~\footnote{https://github.com/lhao499/HDGE}.

\subsection{Image classification}
\label{sec:image_clc}

We compare \methodname{} with
\begin{itemize}
\item The supervised learning baseline uses the standard cross-entropy loss.  We follow the settings of~\citet{zagoruyko2016wide} for evaluation on CIFAR-10 and CIFAR-100, and we decay the learning rate by 0.2 at epoch 60, 120, 160.
\item Supervised Contrastive Learning from~\citep{khosla2020supervised}, which proposes to use label information to select negative examples at the contrastive pre-training stage, and shows incorporating the label information helps the downstream supervised training of classifiers.
We adapt the official implementation of the Supervised Contrastive Loss\footnote{https://github.com/HobbitLong/SupContrast} to use WideResNet.
\item JEM from~\citep{grathwohl2019your}, which proposes to incorporate energy-based modeling training with the standard cross-entropy loss.
\end{itemize}

As reported in Table~\ref{tab:benchmark}, \methodname{} outperforms standard Supervised Learning (which uses only the $q_\theta(y|x)$ loss term), outperforms Supervised Contrastive Learning from~\citet{khosla2020supervised} (which uses a different approximation to the $q_\theta(y|x)$), outperforms JEM (which uses the classification loss on $q_\theta(y|x)$ supplemented with a loss on the marginal $q_\theta(x)$), and outperforms  \methodname{} with $\log q_\theta(x|y)$ (which only trains the generative loss term). This shows the benefit of hybrid discriminative and generative model via jointly optimizing the discriminative (classifier) loss and the generative (contrastive) loss.

In addition, when studying methods that only have the generative term $q_\theta(x|y)$, we see that \methodname{} ($\log q_\theta(x|y)$ only) achieves higher accuracy than~\citet{khosla2020supervised}, which indicates our method provides an improved generative loss term.
\begin{table}[!htbp]
\centering
 \begin{tabular}{c|ccc>{\columncolor[gray]{0.95}}c>{\columncolor[gray]{0.95}}c }
\toprule
 & \multicolumn{5}{c}{Method} \\
 Dataset & \makecell{Supervised \\ Learning} &  \makecell{Supervised\\Contrastive} & JEM &\makecell{\methodname{} (ours)} & \makecell{\methodname{} \\ ($\log q_\theta(x|y)$ only)} \\
 \midrule
 \multirow{1}{*}{CIFAR10} 
 & 95.8 & 96.3 & 94.4 & \bf 96.7 & 96.4 \\
 \midrule
 \multirow{1}{*}{CIFAR100} 
 & 79.9 & 80.5 & 78.1 & \bf 80.9 & 80.6 \\
 \bottomrule
\end{tabular}
\vspace{0.2em}
\caption{\textbf{Comparison on three standard image classification datasets}:
All models use the same batch size of 256 and step-wise learning rate decay, the number of training epochs is 200.
The baselines Supervised Contrastive~\citep{khosla2020supervised}, JEM~\citep{grathwohl2019your}, and our method \methodname{} are based on WideResNet-28-10~\citep{zagoruyko2016wide}.
}
\label{tab:benchmark}
\end{table}

\subsection{Out-of-distribution detection}
\label{sec:ood_exp}

We conduct experiments to evaluate \methodname{} on out-of-distribution (OOD) detection tasks. 
In general, OOD detection is a binary classification problem, where the model is required to produce a score $$s_\theta(x) \in \mathbb{R},$$ where $x$ is the query, and $\theta$ is the model parameters.
We desire that the scores for in-distribution examples are higher than that out-of-distribution examples. 
Following the setting of~\citet{grathwohl2019your}, we use the area under the receiver-operating curve (AUROC)~\citep{hendrycks2016baseline} as the evaluation metric. In our evaluation, we will consider two different score functions, the input density $p(x)$ (\Secref{sec:input_density}) and the predictive distribution $p(y|x)$ (\Secref{sec:pred_dist})

\subsubsection{Input density $p(x)$}
\label{sec:input_density}
Prior work show that fitting a density model on the data and consider examples with low likelihood to be OOD is effective, and the likelihoods from EBMs can be reliably used as a predictor for OOD inputs~\citep{du2019implicit, grathwohl2019your}. We are interested in whether \methodname{} results in better likelihood function for OOD detection.
All the methods are based on the WideResNet-28-10~\citep{zagoruyko2016wide}.
We follow the same experiment settings of~\citet{grathwohl2019your} to remove the batch normalization (BN)~\citep{ioffe2015batch} in WideResNet-28-10.

\begin{figure}[!htbp]
\centering
\begin{tabular}{c||cccc}
\textcolor{blue}{\methodname{}}
&\includegraphics[valign=m,scale=0.25,width=0.2\textwidth]{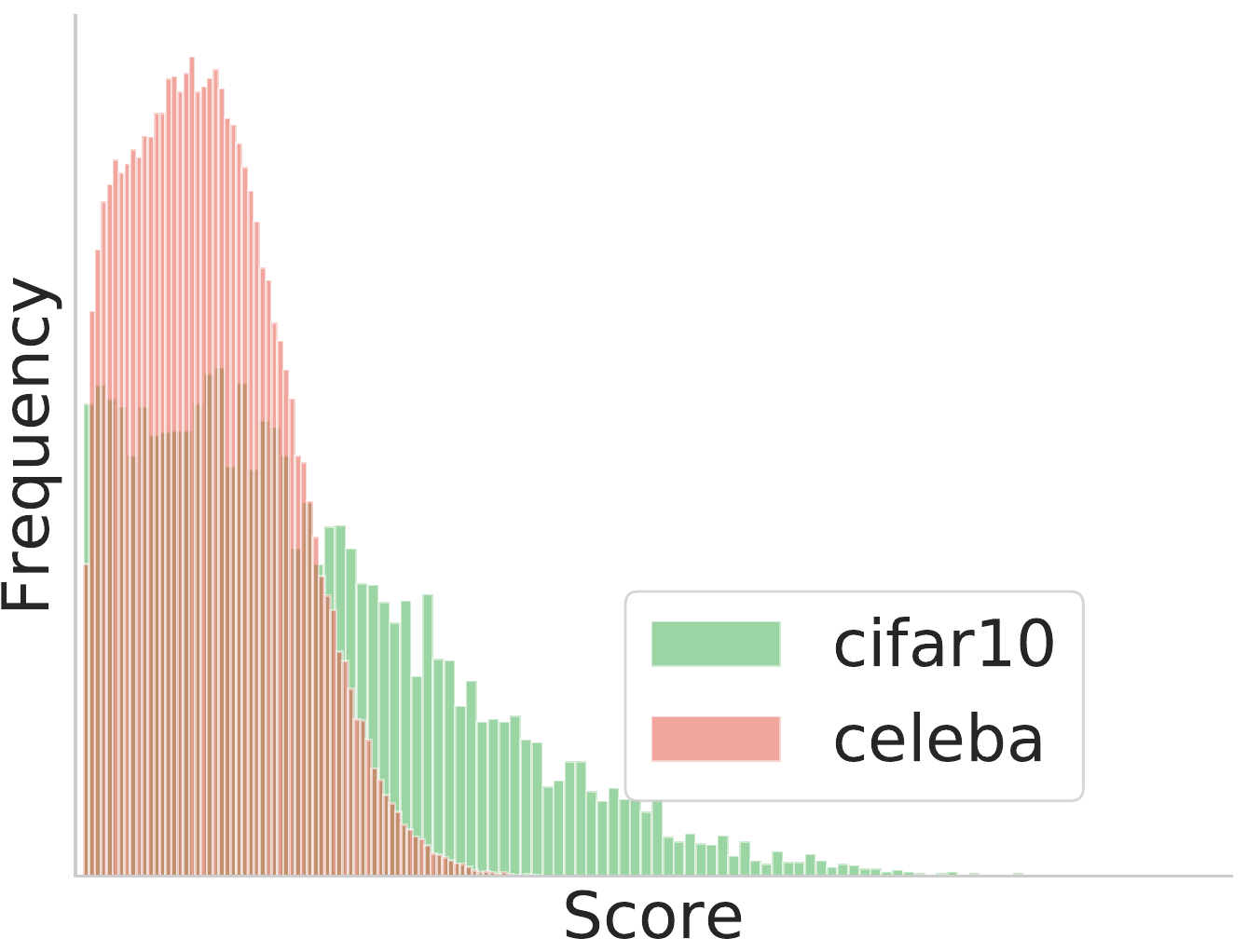}
&\includegraphics[valign=m,scale=0.25,width=0.2\textwidth]{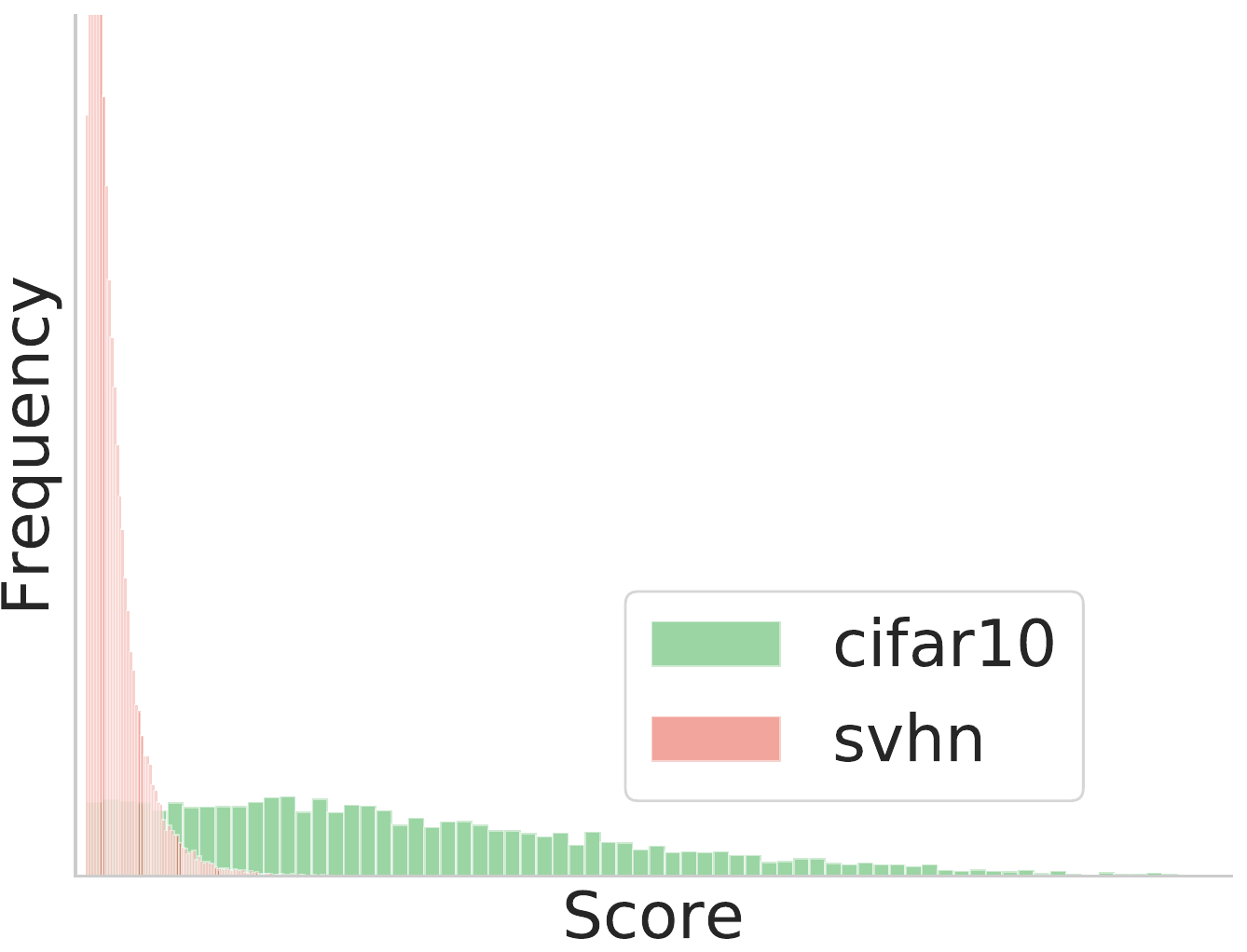}
&\includegraphics[valign=m,scale=0.25,width=0.2\textwidth]{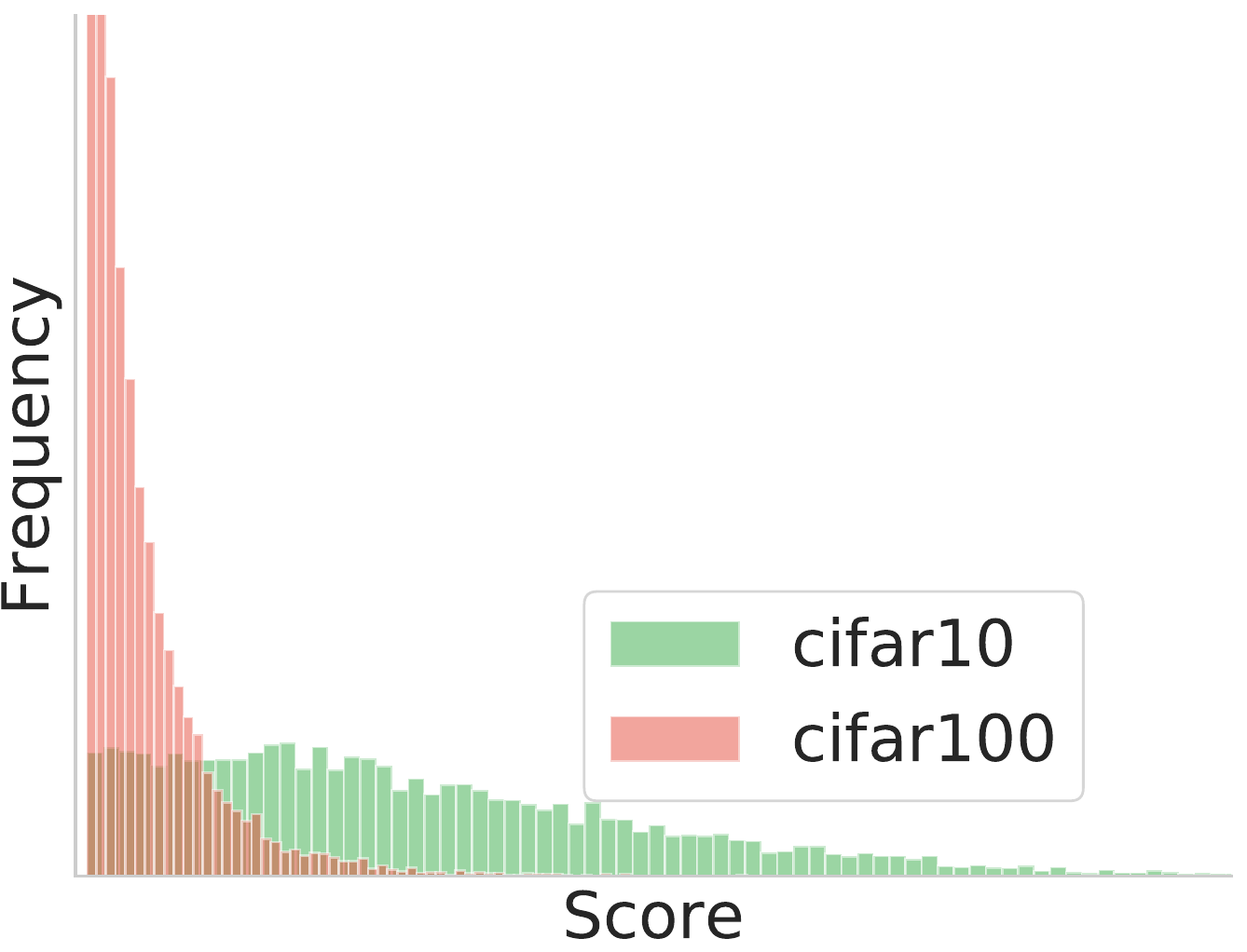}
&\includegraphics[valign=m,scale=0.25,width=0.2\textwidth]{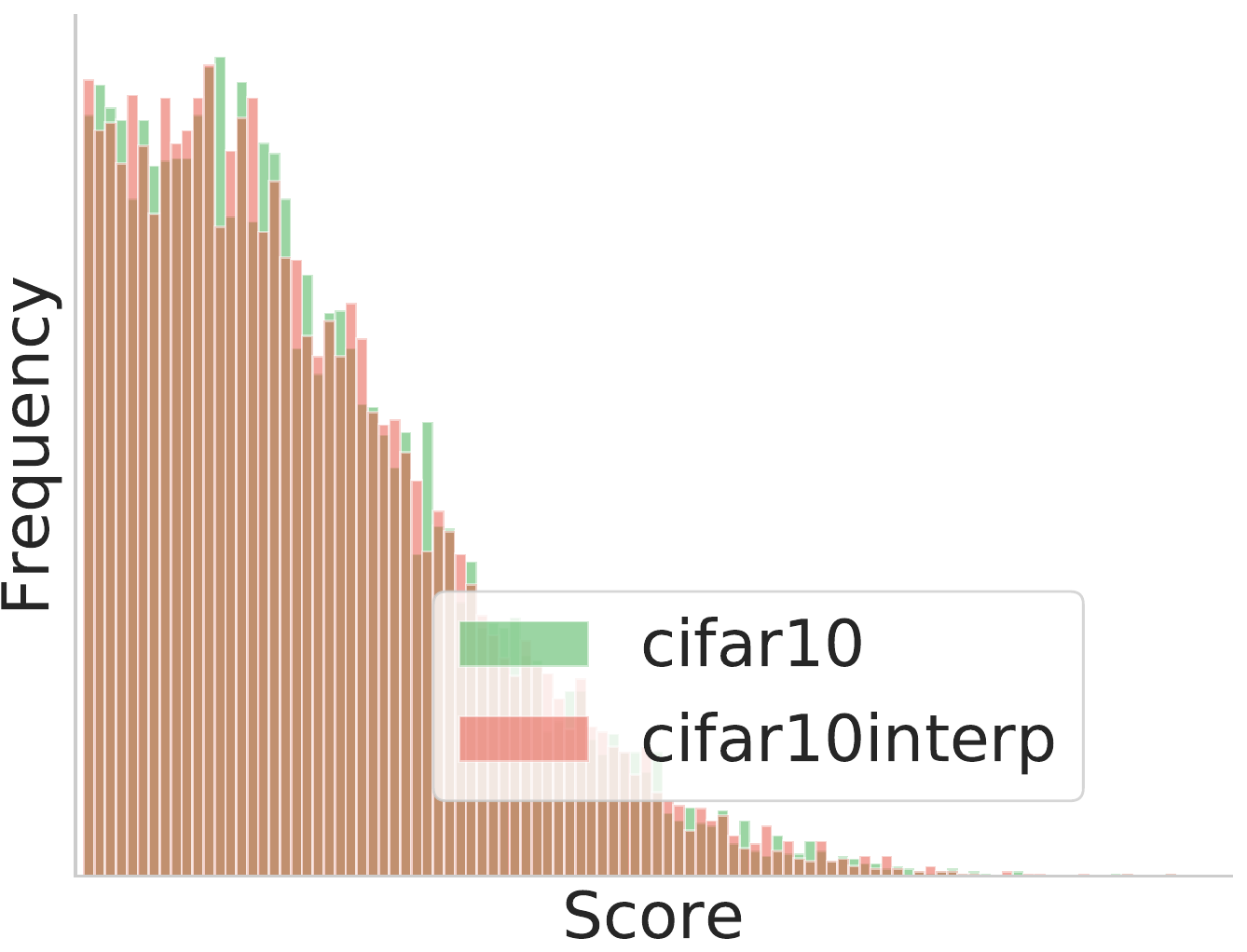}
\\
\textcolor{blue}{JEM}
&\includegraphics[valign=m,scale=0.25,width=0.2\textwidth]{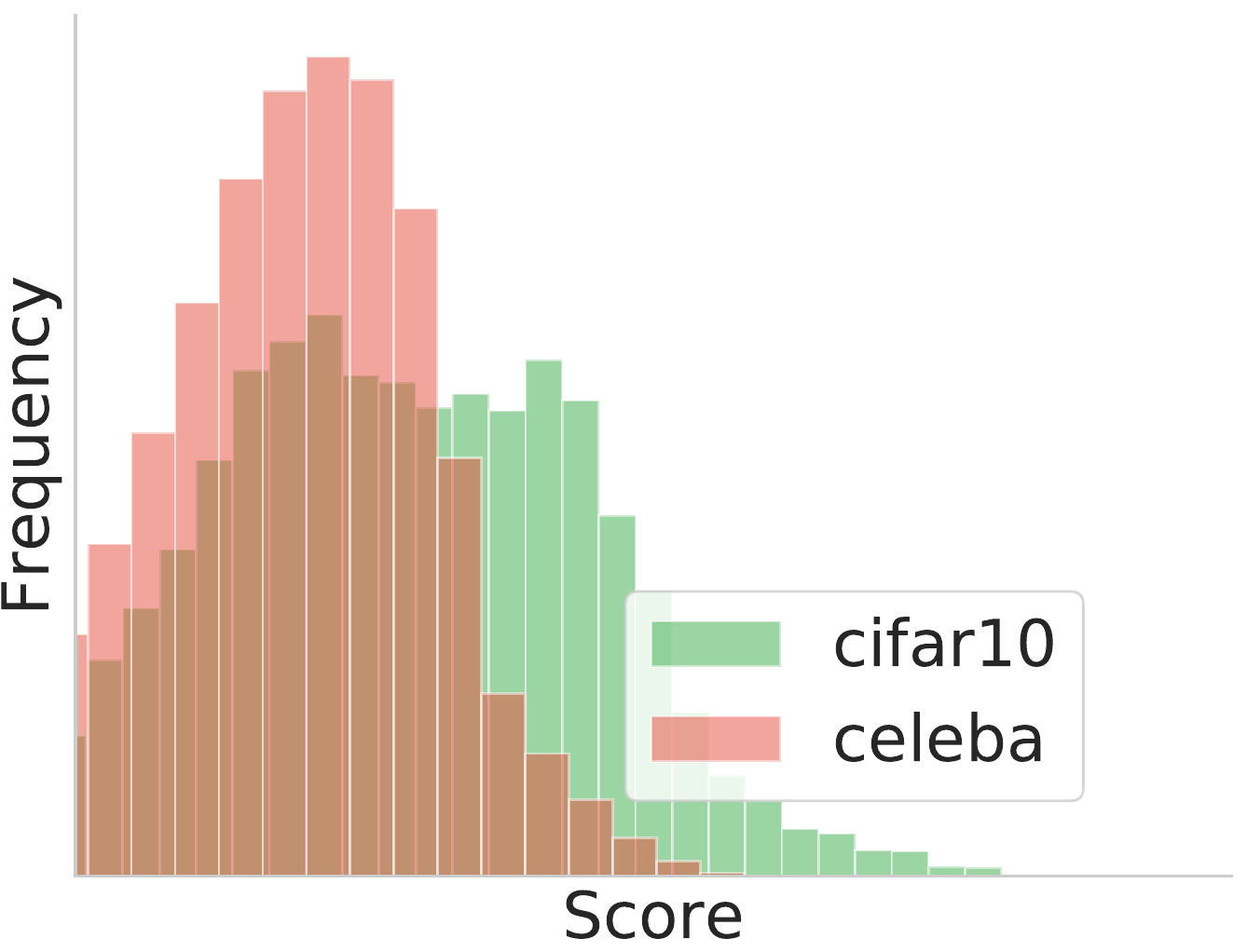}
&\includegraphics[valign=m,scale=0.25,width=0.2\textwidth]{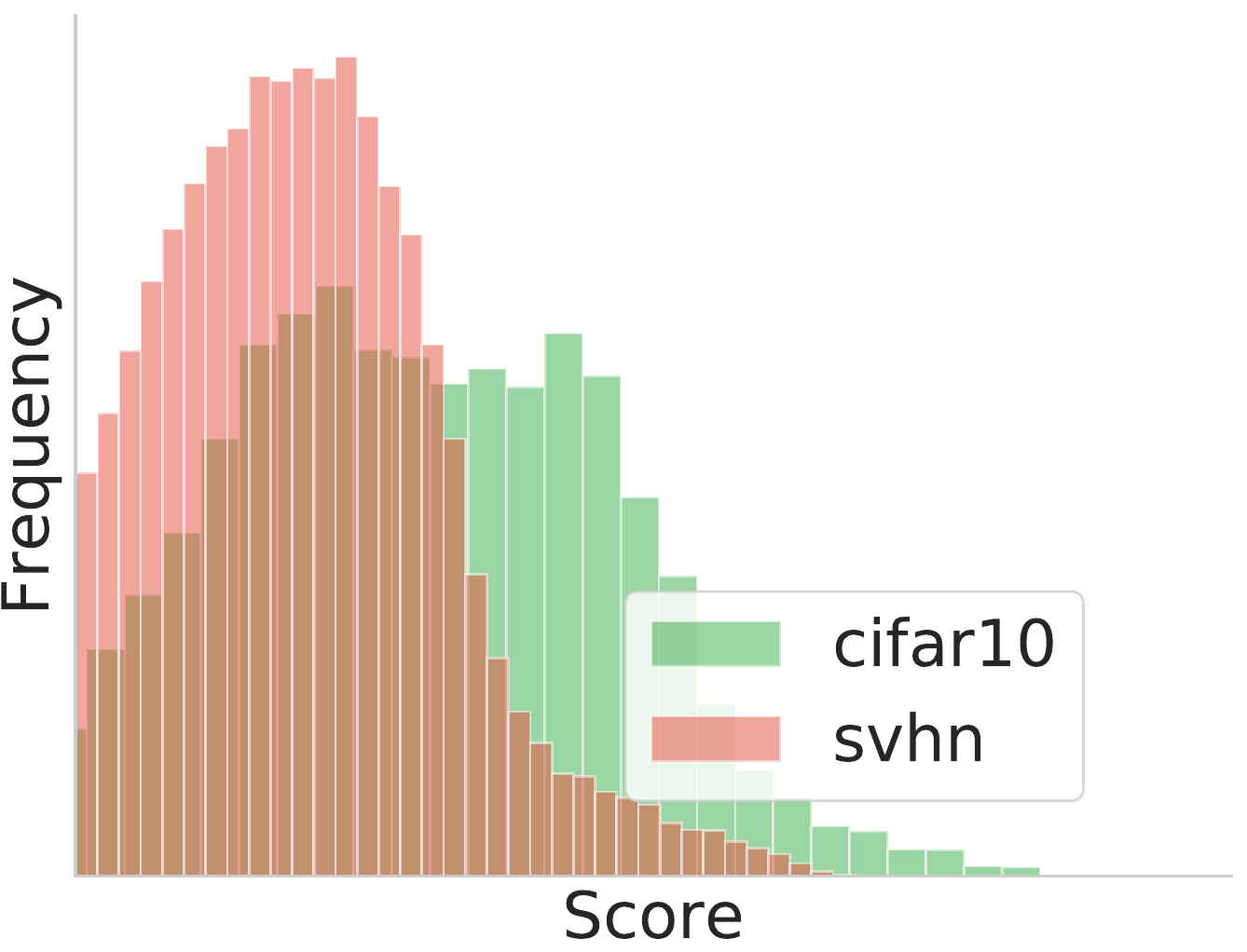}
&\includegraphics[valign=m,scale=0.25,width=0.2\textwidth]{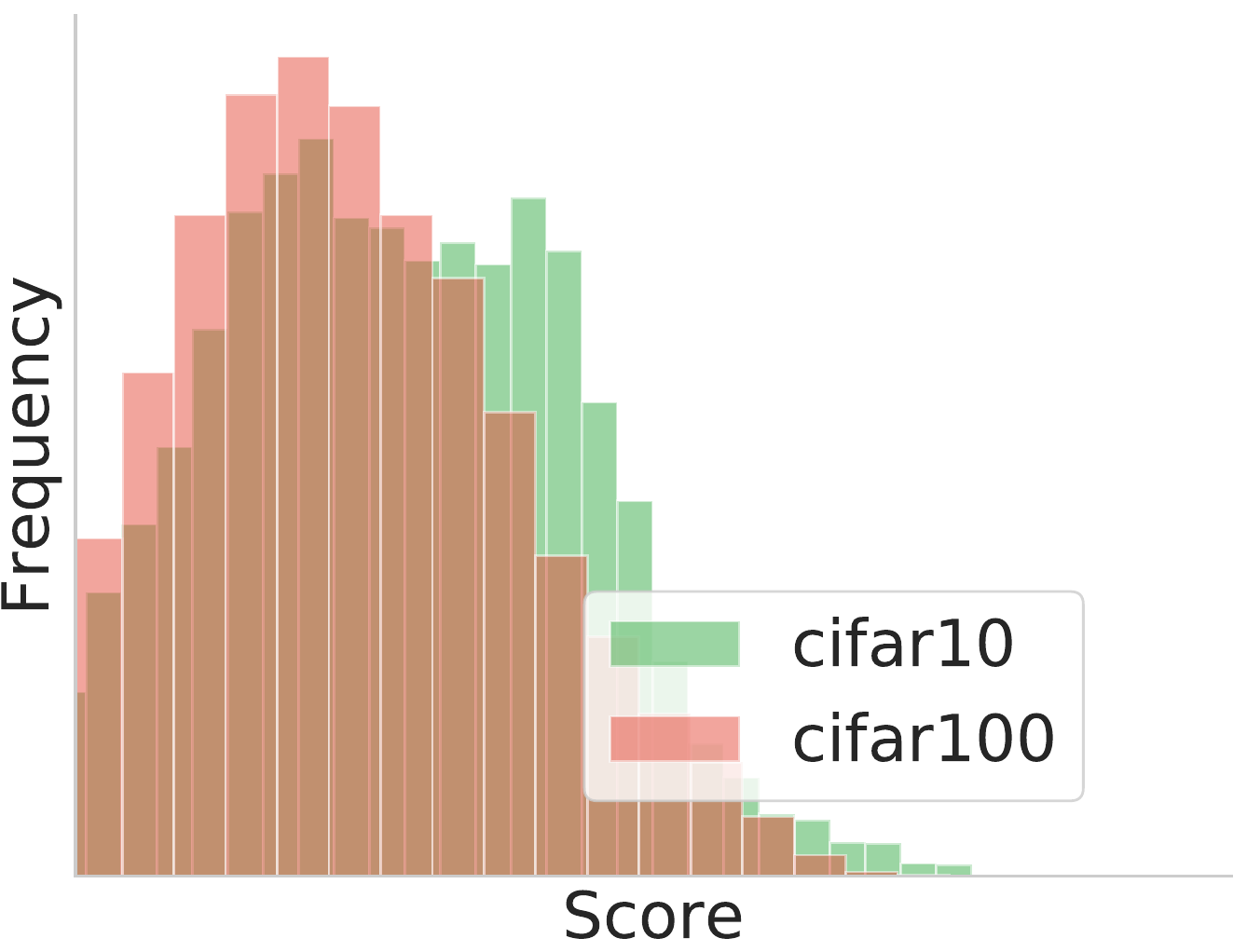}
&\includegraphics[valign=m,scale=0.25,width=0.2\textwidth]{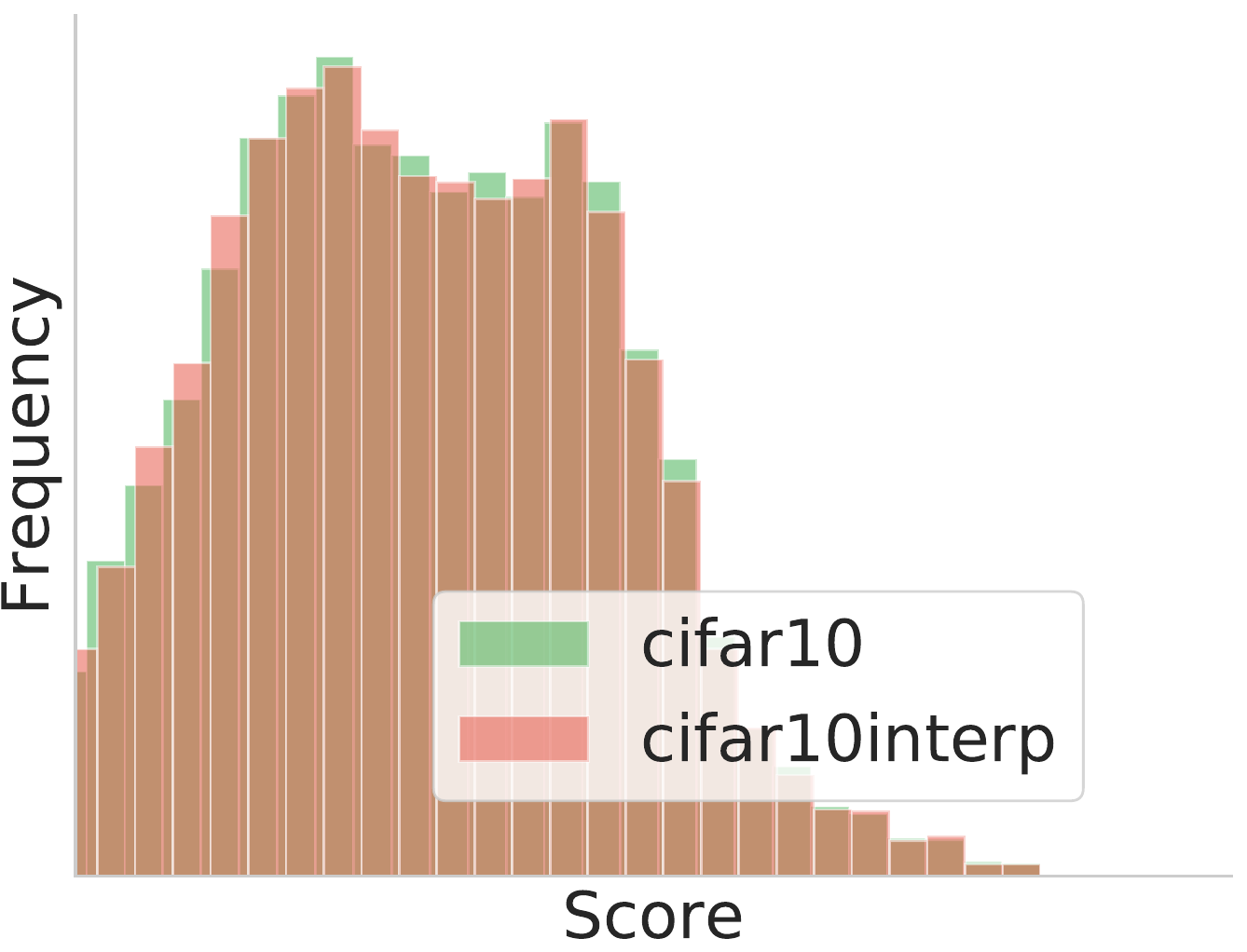}
\\
\end{tabular}
\caption{\textbf{Histograms for OOD detection} using density $p(x)$ as score function. The model is WideResNet-28-10 (without BN). Green corresponds to the score on (in-distribution) training dataset CIFAR-10, and red corresponds to the score on the testing dataset. The cifar10interp denotes a dataset that consists of a linear interpolation of the CIFAR-10 dataset.
}
\label{fig:ood_hist}
\end{figure}

In addition to standard discriminative models and hybrid model JEM, we also compare \methodname{} with
\begin{itemize}
    \item Glow~\citep{kingma2018glow} which is a compelling flow-based generative model.
    \item IGEBM~\citep{du2019implicit} which proposes a method for training EBMs for high dimensional data, and shows training EBMs improves likelihood-based OOD detection.
\end{itemize}
The results are shown in~\Tabref{tab:OODtab} (top), \methodname{} consistently outperforms all of the baselines.
The corresponding distribution of score are visualized in~\Figref{fig:ood_hist}, it shows that \methodname{} correctly assign lower scores to out-of-distribution samples and performs extremely well on detecting samples from SVHN, CIFAR-100, and CelebA. 

We believe that the improvement of \methodname{} over JEM is due to compared with SGLD sampling based methods, \methodname{} holds the ability to incorporate a large number and diverse samples and their corresponding labels information to train the generative conditional $\log p(x|y)$.

\begin{table}[!htbp]
\centering
 \begin{tabular}{ c |c | >{\centering}p{1.5cm} >{\centering}p{1.5cm} >{\centering}p{1.5cm} p{1.5cm}<{\centering} } 
\toprule
 && \multicolumn{4}{c}{Out-of-distribution} \\
 $s_\theta(x)$ & Model & SVHN  & Interp & CIFAR100 & CelebA\\
 \midrule
 \multirow{6}{*}{$\log p(x)$} 
 &WideResNet-28-10 & .46  & .41 & .47 & .49  \\
 &Unconditional Glow & .05 & .51 & .55 & .57 \\
 &Class-Conditional Glow & .07 & .45 & .51 & .53 \\
 &IGEBM & .63 & .70 & .50 & .70 \\
 &JEM & .67 & .65 & .67 & .75 \\
 &\cellcolor{GrayBG} \methodname{} (ours) & \cellcolor{GrayBG} $\bf.96$ & \cellcolor{GrayBG} $\bf.82$ & \cellcolor{GrayBG} $\bf.91$ & \cellcolor{GrayBG} $\bf.80$ \\
 \midrule
 \multirow{5}{*}{$\max_y p(y|x)$} 
 &WideResNet-28-10 & .93  & $\bf.77$ & .85 & .62  \\
 &Contrastive pretraining & .87 & .65 & .80 & .58 \\
 &Class-Conditional Glow & .64 & .61 & .65 & .54  \\
 &IGEBM & .43 & .69 & .54 & .69 \\
 &JEM & .89 & .75 & $\bf.87$ & .79 \\
 & \cellcolor{GrayBG} \methodname{} (ours) & \cellcolor{GrayBG} $\bf.95$ & \cellcolor{GrayBG} $.76$ & \cellcolor{GrayBG} $.84$ & \cellcolor{GrayBG} $\bf.81$ \\
 \bottomrule
\end{tabular}
\caption{\textbf{OOD Detection Results}. The model is WideResNet-28-10 (without BN). The training dataset is CIFAR-10. Values are AUROC. Results of the baselines are from~\citet{grathwohl2019your}.}
\label{tab:OODtab}
\end{table}

\subsubsection{Predictive Distribution $p(y|x)$}
\label{sec:pred_dist}
A widely used OOD score function is the maximum prediction probability~\citep{hendrycks2016baseline} which is given by $s_\theta(x) = \max_{y} p_\theta(y|x)$. 
Intuitively, a model with high classification accuracy tends to has a better OOD performance using this score function.

We compare with \methodname{} with standard discriminative models, generative models, and hybrid models. We also evaluate a contrastive pre-training baseline which consists of learning a representation via contrastive learning and training a linear classifier on top of the representation.

The results of OOD detection are show in~\Tabref{sec:ood_exp} (bottom). 
We find \methodname{} performs beyond the performance of a strong baseline classifier and considerably outperforms all other generative modeling and hybrid modeling methods.

The OOD detection evaluation shows that it is helpful to jointly train the generative model $p(x|y)$ together with the classifier $p(y|x)$ to have a better classifier model. \methodname{} provides an effective and simple approach to improve out-of-distribution detection.

\subsection{Calibration}
\label{sec:cal_exp}
Calibration plays an important role when deploy the model in real-world scenarios where outputting an incorrect decision can have catastrophic consequences~\citep{guo2017calibration}. 
The goodness of calibration is usually evaluated in terms of the Expected Calibration Error (ECE), which is a metric to measure the calibration of a classifier. 

It works by first computing the confidence, $\max_y p(y|x_i)$, for each $x_i$ in some dataset and then grouping the items into equally spaced buckets $\{B_m\}_{m=1}^M$ based on the classifier's output confidence. 
For example, if $M=20$, then $B_0$ would represent all examples for which the classifier's confidence was between $0.0$ and $0.05$. 
The ECE is defined as following:
\begin{align}
    \text{ECE} = \sum_{m=1}^M \frac{|B_m|}{n} |\text{acc}(B_m) - \text{conf}(B_m) |,
\end{align}
where $n$ is the number of examples in the dataset, acc$(B_m)$ is the averaged accuracy of the classifier of all examples in $B_m$ and conf$(B_m)$ is the averaged confidence over all examples in $B_m$.
For a perfectly calibrated classifier, this value will be 0 for any choice of $M$.  
Following~\citet{grathwohl2019your}, we choose $M=20$ throughout the experiments.
A classifier is considered calibrated if its predictive confidence, $\max_y p(y|x)$, aligns with its misclassification rate. Thus, when a calibrated classifier predicts label $y$ with confidence score that is the same at the accuracy.

\begin{figure}[!htbp]
\centering
\begin{tabular}{ccc}
 \includegraphics[width=0.3\textwidth]{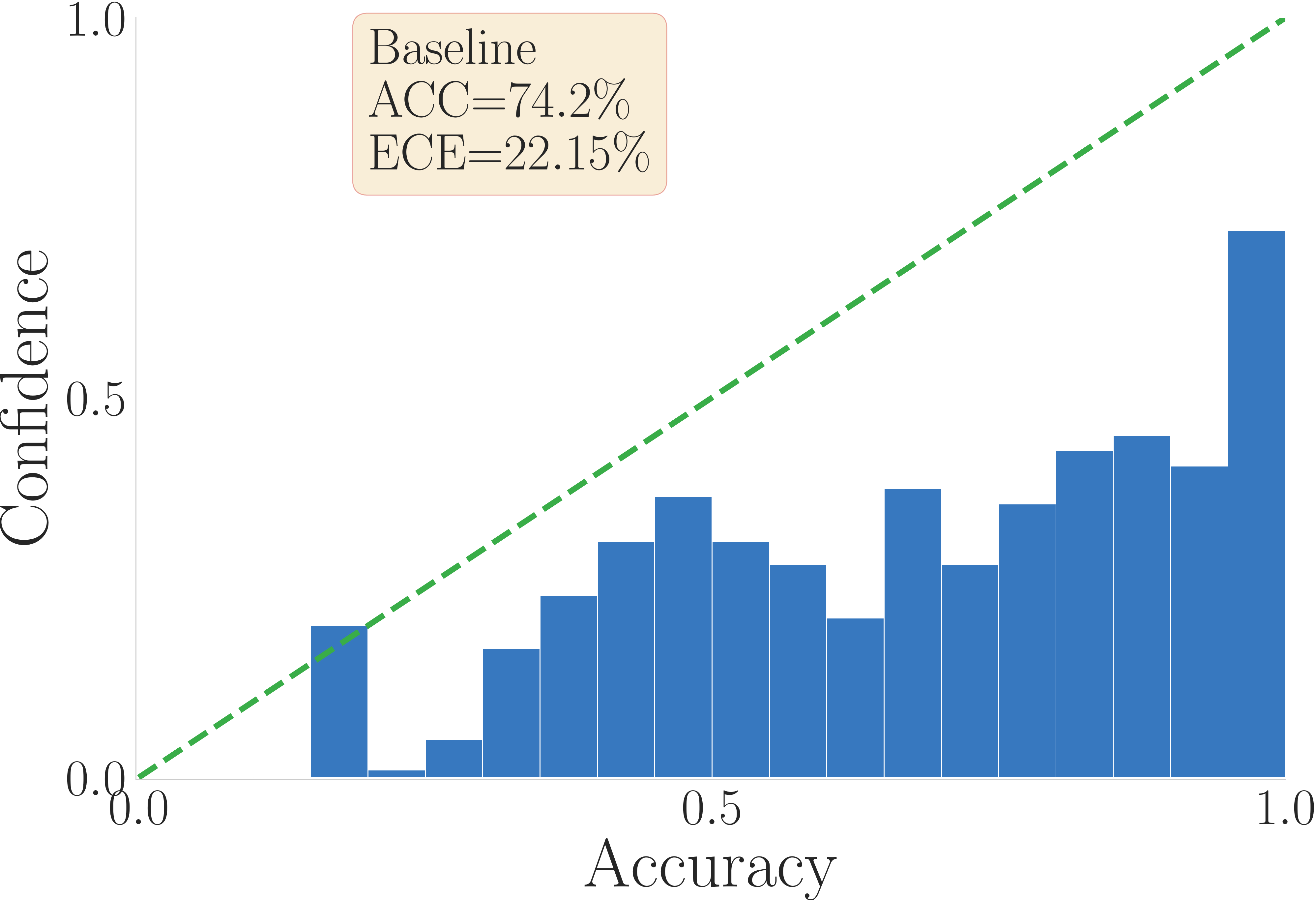}
&\includegraphics[width=0.3\textwidth]{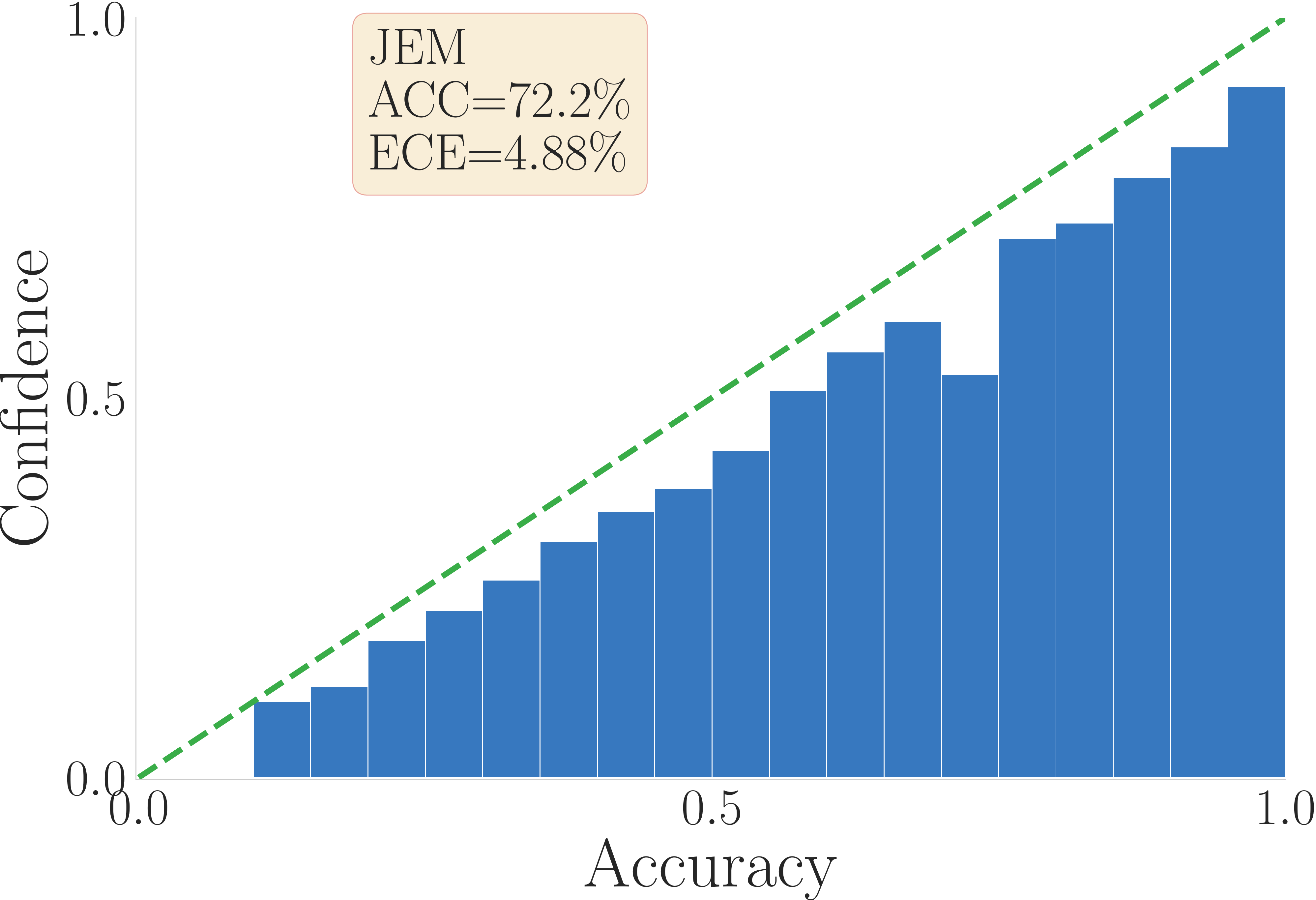}
&\includegraphics[width=0.3\textwidth]{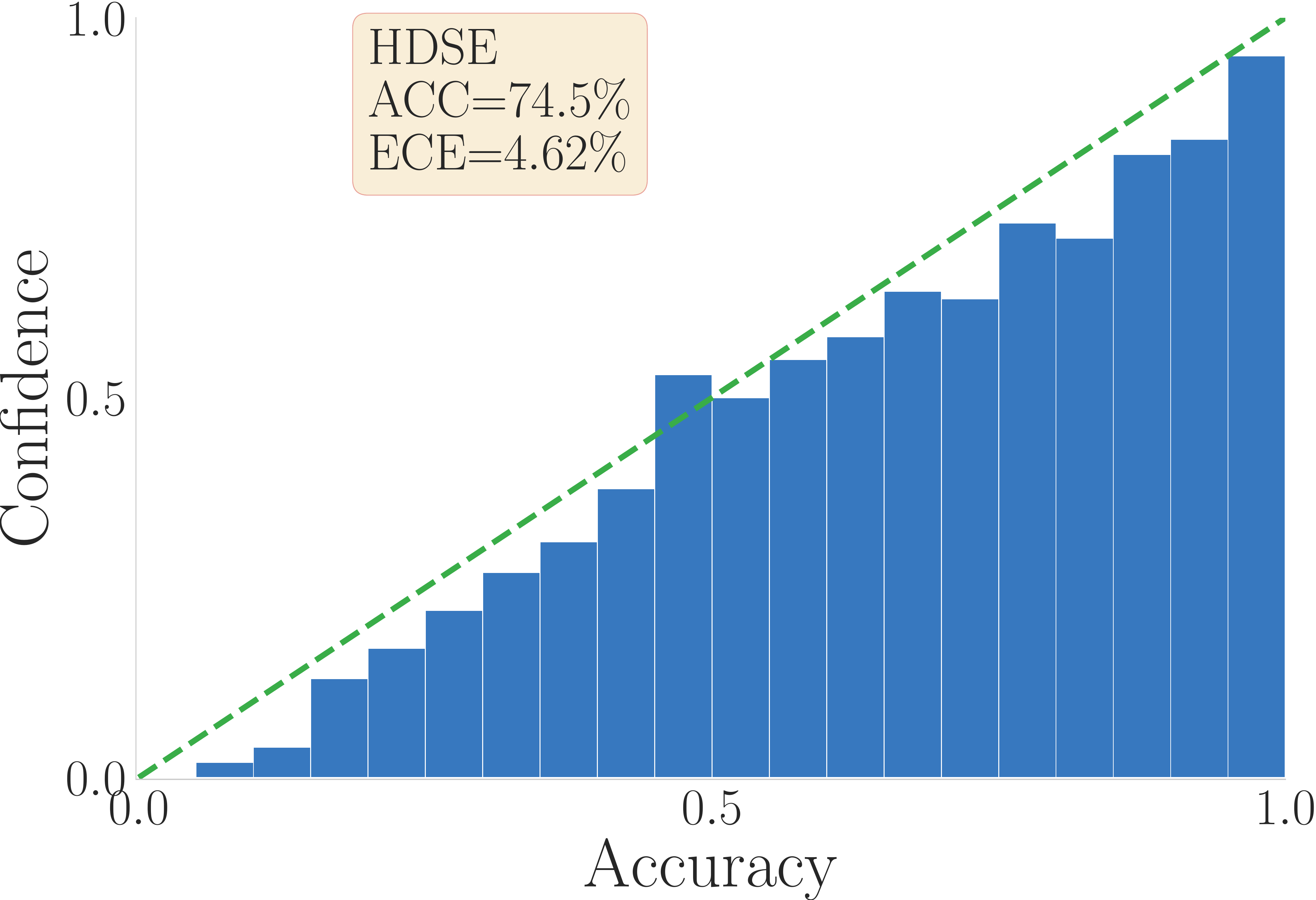}\\
 \end{tabular}
\caption{\textbf{CIFAR-100 calibration results}. The model is WideResNet-28-10 (without BN). Expected calibration error (ECE)~\citep{guo2017calibration} on CIFAR-100 dataset under various training losses.}
\label{fig:cal}
\end{figure}

We evaluate the methods on CIFAR-100 where we train \methodname{} and baselines of the same architecture, and compute the ECE on hold-out datasets.
The histograms of confidence and accuracy of each method are shown in~\Figref{fig:cal}. 

While classifiers have grown more accurate in recent years, they have also grown considerably less calibrated~\citep{guo2017calibration}, as shown in the left of~\Figref{fig:cal}.
\citet{grathwohl2019your} significantly improves the calibration of classifiers by optimizing $p(x)$ as EBMs training (\Figref{fig:cal} middle), however, their method is computational expensive due to the contrastive divergence and SGLD sampling process and their training also sacrifices the accuracy of the classifiers. In contrast, \methodname{} provides a computational feasible method to significantly improve both the accuracy and the calibration at the same time (\Figref{fig:cal} right).

\subsection{Adversarial robustness}
\label{sec:adv_robust}
\begin{wrapfigure}[22]{R}{0.3\textwidth}
\centering
\begin{tabular}{c}
\includegraphics[width=0.3\textwidth]{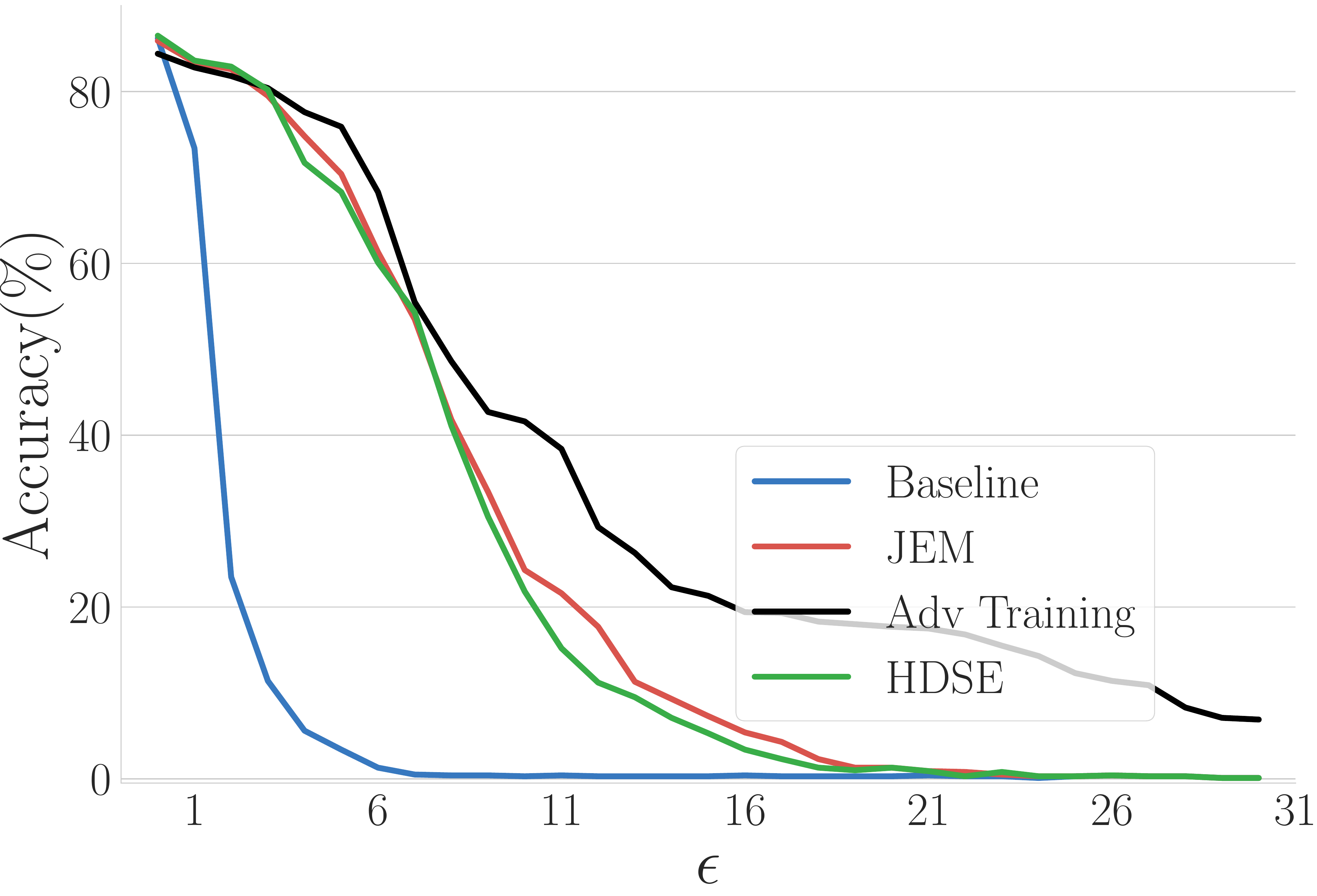} \\ 
(a) $L_\infty$ Robustness\\
\includegraphics[width=0.3\textwidth]{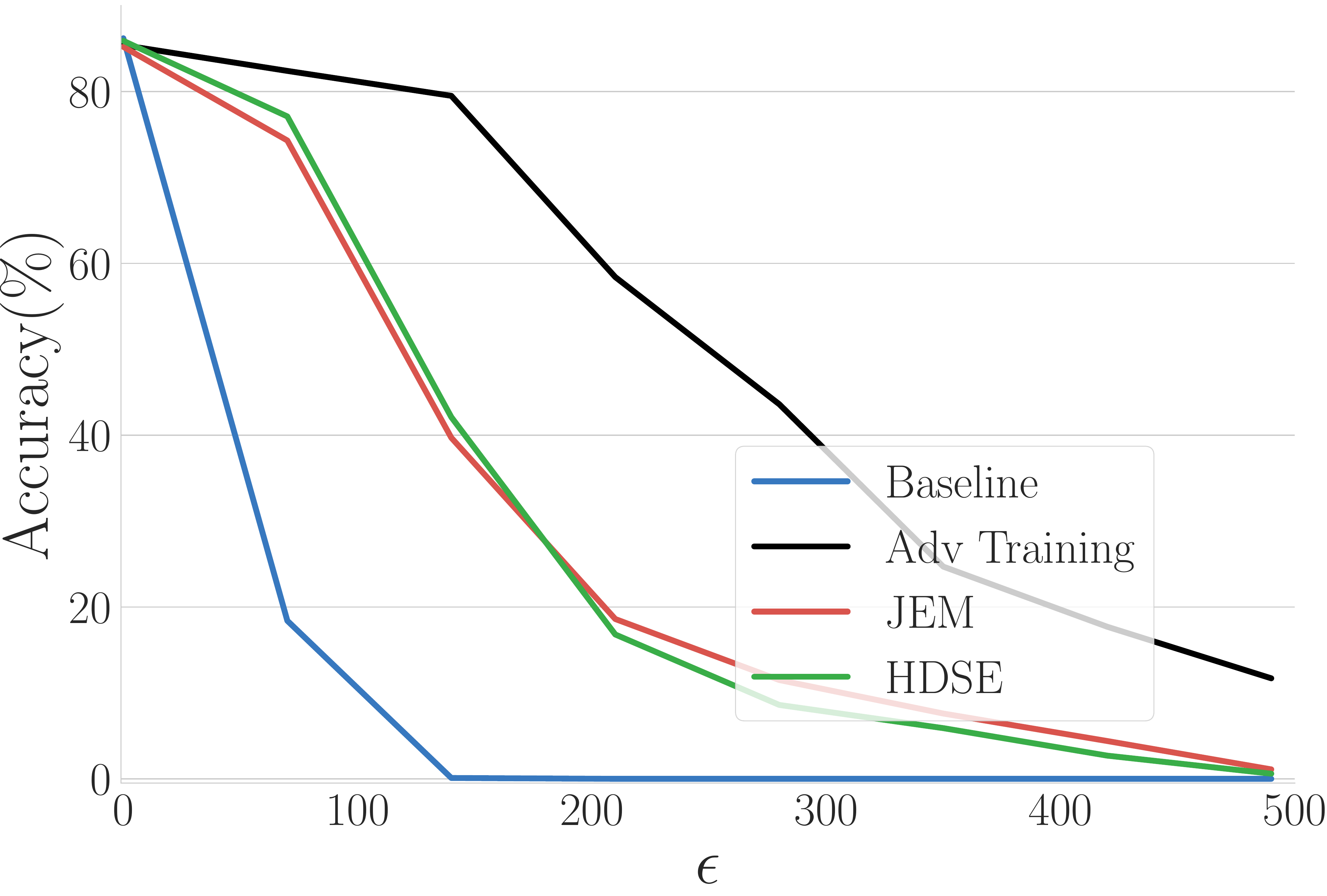} \\
(b) $L_2$ Robustness
\end{tabular}
\caption{\textbf{Adversarial robustness} results with PGD attacks. 
\methodname{} adds considerable robustness to standard supervised training and achieves comparable robustness with JEM.}
\label{fig:pgd-main}
\end{wrapfigure}
While image classifiers have seen many progress in accuracy, they are known to be sensitive to perturbation-based adversarial examples on input~\citep{goodfellow2014explaining}.
These examples exploit the model's sensitivity to semantically meaningless perturbations: while examples are closed to real inputs in datasets in terms of a given metric, they don not necessarily reside within areas of high density according to the model distribution, therefore, the model can wrongly classify these rarely encountered inputs.
In order to deploy the classification systems to real-world applications, such as pedestrian detection and traffic sign recognition on autonomous cars, it is extremely important to develop models that are robust to adversarial examples.

The commonly considered adversarial attack is the $L_p$-norm constrained adversarial examples, which are defined as $\hat{x} \in B(x, \epsilon)$ that changes the model's prediction, where $B(x, r)$ denotes a ball centered at $x$ with radius $r$ under the $L_p$-norm metric. In this work, we run white-box PGD (projected gradient descent) attack with respect to the $L_2$ and $L_\infty$ norms, giving the attacker access to gradients, in which PGD is used to find a local maximal within a given perturbation ball~\citep{madry2017towards}.
We train \methodname{} and compare with the state-of-the-art adversarial training methods. 
\begin{itemize}
    \item Adv Training~\citep{madry2017towards,santurkar2019image} which proposes to use robust optimization to train classifier to be robust to the norm through which it is being attacked.
\end{itemize}
Results from the PGD experiments can be seen in~\Figref{fig:pgd-main}. 
We can see that \methodname{} can achieve compelling robustness to the state-of-the-art adversarial training methods. 

We note that while JEM improves the robustness too by optimizing the likelihood of EBMs, it requires computationally expensive SGLD sampling procedure. In contrast, \methodname{} significantly improves the robustness of standard classifiers by computationally scalable contrastive learning.

\subsection{Goodness of approximation}
\label{sec:good_approximation}
\begin{wrapfigure}[13]{R}{0.3\textwidth}
\centering
\begin{tabular}{c}
\includegraphics[width=.3\textwidth]{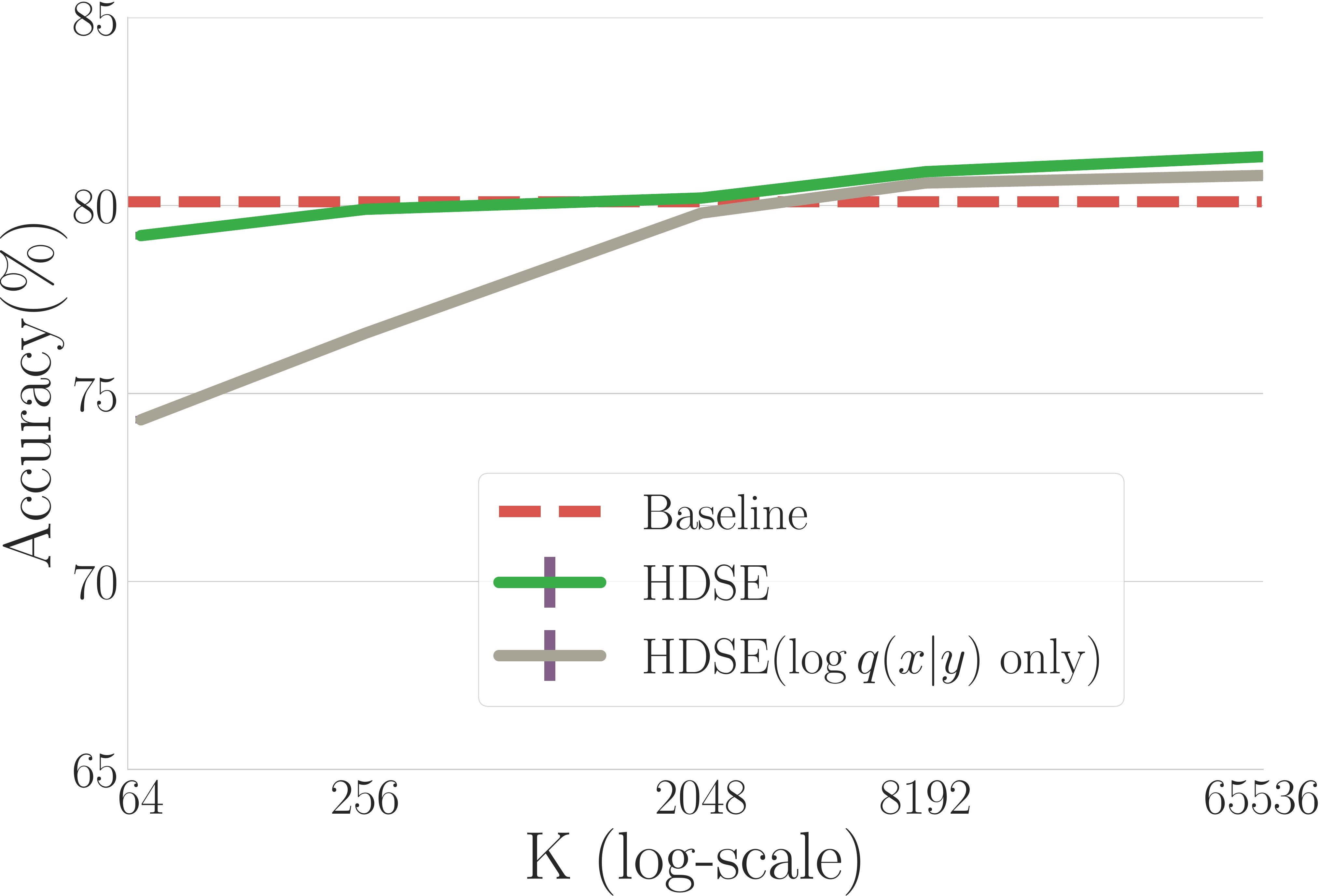} \\
\end{tabular}
\caption{
{\textbf{Accuracy comparison} with respect to different $K$ on CIFAR-100.
The baseline is standard cross-entropy loss.
The model is WideResNet-28-10. Batch size is 256.
} 
}
\label{fig:acc_k}
\end{wrapfigure}
Since we made the approximation to energy-based model by contrastive learning in~\Eqref{eq:k_approx}, we are interested in evaluating the impact of the number of negative examples $K$ on the goodness of this approximation. 
We consider a classification task and a density based OOD detection task as proxies of evaluating the approximation. 
\paragraph{Classification.}
We compare the image classification of \methodname{} on CIFAR-100. The results are shown in~\Figref{fig:acc_k}. We found that increasing number of negative samples $K$ improves the performance of~\methodname{}, and with sufficient number of negative examples~\methodname{} significantly outperform the cross-entropy loss. 
The reason may be training with many negative examples helps to discriminate between positive and negative samples.

\paragraph{OOD detection.}
We evaluate \methodname{} with different value of $K$ by running experiments on the $\log p(x)$ based OOD tasks, we use the same experiments setting as~\Secref{sec:ood_exp}.

We vary the batch size of SGLD sampling process in~\citet{grathwohl2019your}, effectively, we change the number of samples used to estimate the derivative of the normalization constant $\E_{p_\theta(x')}\left[ \frac{\partial E_\theta(x')}{\partial \theta} \right]$ in the JEM update rule~\Eqref{eq:ebm_grad_est}.
Specifically, we increase the default batch size $N$ from 64 to 128 and 256, due to running the SGLD process is memory intensive and the technique constraints of the limited CUDA memory, we were unable to further increase the batch size.
We also decrease $K$ in \methodname{} to $\{64, 128, 256\}$ to study the effect of approximation.

The results are shown in~\Tabref{tab:ood_k}, the results show that \methodname{} with a small $K$ performs fairly well except on CelebA probably due to the simplicity of other datasets. We note \methodname{}($K = 64$) outperforms JEM and three out of four datasets, which shows the approximation in \methodname{} is reasonable good. 
While increasing batch size of JEM improves the performance, we found increasing $K$ in \methodname{} can more significantly boost the performance on all of the four datasets.
We note JEM with a large batch size is significantly more computational expensive than \methodname{}, as a result JEM runs more slower than \methodname{} with the largest $K$.
\begin{table}[!htbp]
\centering
 \begin{tabular}{ c |c | >{\centering}p{1.5cm} >{\centering}p{1.5cm} >{\centering}p{1.5cm} p{1.5cm}<{\centering} } 
\toprule
 && \multicolumn{4}{c}{Out-of-distribution} \\
 $s_\theta(x)$ & Model & SVHN  & Interp & CIFAR100 & CelebA\\
 \midrule
 \multirow{7}{*}{$\log p(x)$} 
 &JEM ($N=64$) (default) & .67 &  .65 & .67 & .75 \\
 &JEM ($N=128$) & .69& .67& .68& .75\\
 &JEM ($N=256$) & .70& .69& .68& .76\\
 & \cellcolor{GrayBG} \methodname{} ($K=64$)
 & \cellcolor{GrayBG} $\bf.89$ & \cellcolor{GrayBG} $\bf.79$ & \cellcolor{GrayBG} $\bf.84$ & \cellcolor{GrayBG} $\bf.62$ \\
 & \cellcolor{GrayBG} \methodname{} ($K=128$) 
 & \cellcolor{GrayBG} $\bf.91$ & \cellcolor{GrayBG} $\bf.80$ & \cellcolor{GrayBG} $\bf.89$ & \cellcolor{GrayBG} $\bf.73$ \\
 & \cellcolor{GrayBG} \methodname{} ($K=256$) 
 & \cellcolor{GrayBG} $\bf.93$ & \cellcolor{GrayBG} $\bf.81$ & \cellcolor{GrayBG} $\bf.90$ & \cellcolor{GrayBG} $\bf.76$ \\
 & \cellcolor{GrayBG} \methodname{} ($K=65536$) (default) & \cellcolor{GrayBG} $\bf.96$ & \cellcolor{GrayBG} $\bf.82$ & \cellcolor{GrayBG} $\bf.91$ & \cellcolor{GrayBG} $\bf.80$ \\
 \bottomrule
\end{tabular}
\caption{\textbf{Ablation of approximation} on detecting OOD samples. We use CIFAR 10 for in-distribution.
}
\label{tab:ood_k}
\end{table}

\subsection{Generative modeling tasks}
\label{sec:gen_model}

\begin{figure}[!htbp]
\begin{minipage}{\textwidth}
\begin{minipage}[b]{0.55\textwidth}
\centering
\begin{tabular}{|c|c|ccc|}
        \hline
        Class & Model & Accuracy\% $\uparrow$ & IS$\uparrow$ & FID$\downarrow$ \\
        \hline
        \multirow{7}{*}{\textbf{Hybrid}} &  Residual Flow & 70.3 & 3.6 & 46.4\\
        &Glow &  67.6  &  3.92 & 48.9\\
        &IGEBM & 49.1 & 8.3 & 37.9\\
        &JEM & 92.9 & 8.76 & 38.4 \\
        &\cellcolor{GrayBG} \methodname{} & \cellcolor{GrayBG} $\bf{94.6}$ & \cellcolor{GrayBG} N/A & \cellcolor{GrayBG} N/A \\
        &\cellcolor{GrayBG} \methodname{}+JEM & \cellcolor{GrayBG} 94.4 & \cellcolor{GrayBG} $\bf{9.19}$ & \cellcolor{GrayBG} $\bf{37.6}$ \\
        \hline
        \multirow{2}{*}{\textbf{Disc.}} &WideResNet & 95.8 & N/A & N/A \\
        &\makecell{WideResNet(w/o BN)} & 93.6 & N/A & N/A \\
        \hline
        \multirow{2}{*}{\textbf{Gen.}} &SNGAN & N/A & 8.59 & 25.5 \\
        & NCSN & N/A & 8.91 & 25.32\\
        \hline
\end{tabular}
\captionof{table}{\textbf{Hybrid modeling results} on CIFAR-10.
All models are based on WideResNet-28-10~\citep{zagoruyko2016wide}(without BN). 
Residual Flow~\citep{chen2019residual}, Glow~\citep{kingma2018glow}, IGEBM~\citep{du2019implicit}, SNGAN~\citep{miyato2018spectral}, NCSN~\citep{song2019generative}, JEM~\citep{grathwohl2019your}}
\label{tab:gen-tab}
\end{minipage}
\hspace{5em}
\begin{minipage}[b]{0.35\textwidth}
\centering
\includegraphics[height=35.5mm]{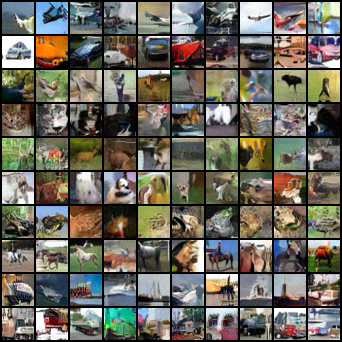}
\captionof{figure}{\textbf{Class-conditional samples} generated by running \methodname{} + JEM on CIFAR-10.}
\label{fig:cifar-samp}
\end{minipage}
\end{minipage}
\end{figure}

\methodname{} models can be sampled from with SGLD.  However, during experiments we found that adding the marginal log-likelihood over $x$ (as done in JEM) improved the generation.   we hypothesis that this is due the approximation via contrastive learning focuses on discriminating between images of different categories rather than estimating density. 

So we evaluated generative modeling through SGLD sampling from a model trained with the following objective:
\begin{align}
    \min_\theta \E_{p_{\rm data}(x, y)} \left[\log q_\theta(y|x) + \log q_\theta(x|y) + \log q_\theta(x) \right], 
\end{align}
where $\log q_\theta(x)$ is optimized by running SGLD sampling and contrastive divergence as in JEM and $\log q_\theta(y|x) + \log q_\theta(x|y)$ is optimized through \methodname{}. 

We train this approach on CIFAR-10 and compare against other hybrid models as well as standalone generative and discriminative models.  We present inception scores (IS)~\citep{salimans2016improved} and Frechet Inception Distance (FID)~\citep{heusel2017gans} given that we cannot compute normalized likelihoods.  The results are shown in~\Tabref{tab:gen-tab} and~\Figref{fig:cifar-samp}.

The results show that jointly optimizing $\log q_\theta(y|x) + \log q_\theta(x|y) + \log q_\theta(x)$ by \methodname{} (first two terms) and JEM (third term) together can outperform optimizing $\log q_\theta(y|x) + \log q_\theta(x)$ by JEM, and it significantly improves the generative performance over the state of the art in generative modeling methods and retains high classification accuracy simultaneously. We believe the superior performance of \methodname{} + JEM is due to the fact that \methodname{} learns a better classifier and JEM can exploit it and maybe optimizing $\log p(x|y)$ via \methodname{} is a good auxiliary objective.

\section{Related work}
Prior work~\citep{khosla2020supervised} connects supervised learning with contrastive learning by using label information to cluster positive and negative examples, and show improved image classification and hyperparameters robustness in contrastive learning. Their work applied to learning a representation for downstream tasks, while this work introduces an end-to-end framework of contrastive learning that leverages label information for supervised learning, and show improved performance on discriminative and generative modeling tasks.

\citet{ng2002discriminative, raina2004classification, lasserre2006principled, larochelle2008classification} compare and study the connections and differences between discriminative model and generative model, and shows hybrid generative discriminative models can outperform purely discriminative models and purely generative models. 
This work differs in that we provide an effective training approach to optimize the hybrid model, motivated from the energy-based models.

\citet{xie2016theory, du2019implicit} demonstrate the EBMs can be derived from classifiers, they reinterpret the logits to define a class-conditional EBM $p(x|y)$,~\citet{grathwohl2019your} shows the alternative class-conditional EBM $p(y|x)$ leads to significant improvement in generative modeling while retain compelling classification accuracy.
Prior work also explicitly adopt discriminative models for generative modelling and show this iterative bootstrapping leads to improvement in classification, texture modeling, and other applications~\citep{tu2007learning, lazarow2017introspective, jin2017introspective}.
Our method \methodname{} is built upon~\citet{grathwohl2019your} which shows a standard classifier are secretly a generative model. 
The likelihood $p(x)$ in~\citet{grathwohl2019your} is trained by using Contrastive Divergence~\citep{tieleman2008training} and SGLD~\citep{welling2011bayesian}, we reveal that the likelihood $p(x)$ can be approximated by contrastive learning, and this novel loss leads to significant improvement in various applications. 
We note optimizing our loss function is more computational efficient than training EBMs on high-dimensional data~\citet{grathwohl2019your} and this training does not suffer from numerical stability issues.
While there are advances in scaling the training of EBMs to to high-dimensional data~\citep{nijkamp2019learning, du2019implicit, grathwohl2019your}, it remains computational expensive and often costs orders magnitude of more time and computation to training EBMs and therefore makes it difficult to scale. On the contrary~\methodname{} holds the promise to scale more gracefully, which makes it a promising direction to apply EBMs to even wider applications.

\section{Conclusion}
In this work, we develop \methodname{}, a new framework for supervised learning and contrastive learning through the perspective of hybrid discriminative and generative model. 
We propose to leverage contrastive learning to approximately optimize the model for discriminative and generative tasks, and effectively take advantage of the large number of samples from different classes.
Our framework provides a simple, scalable and general method for solving supervised and contrastive machine learning problems.

\section{Acknowledgement}
This research was supported by DARPA Data-Driven Discovery of Models (D3M) program. We would like to thank Will Grathwohl, Kuan-Chieh Wang, Kimin Lee, Qiyang Li, Misha Laskin, and Jonathon Ho for improving the presentation and giving constructive comments. We would also like to thank Aravind Srinivas for providing tips on training models. We would also like to thank Kimin Lee for helpful discussions on OOD detection and robustness tasks.

\bibliography{Mathbib/master}
\bibliographystyle{neurips}

\end{document}